\documentclass[nohyperref]{article}

\usepackage{microtype}
\usepackage{graphicx}
\usepackage{subfigure}
\usepackage{booktabs} %

\usepackage[hyphens]{xurl}
\usepackage{hyperref}
\usepackage[hyphenbreaks]{breakurl}

\usepackage[accepted]{icml2023}

\usepackage{amsmath}
\usepackage{amssymb}
\usepackage{mathtools}
\usepackage{amsthm}

\usepackage[capitalize,noabbrev]{cleveref}

\theoremstyle{plain}

\theoremstyle{definition}

\theoremstyle{remark}

\usepackage[textsize=tiny]{todonotes}

\icmltitlerunning{%
Set--Membership Inference Attacks using Data Watermarking.}

\begin{document}

\twocolumn[
\icmltitle{%
Set--Membership Inference Attacks using Data Watermarking}

\icmlsetsymbol{equal}{*}

\begin{icmlauthorlist}
\icmlauthor{Mike Laszkiewicz}{equal,math,cs}
\icmlauthor{Denis Lukovnikov}{equal,cs}
\icmlauthor{Johannes Lederer}{math}
\icmlauthor{Asja Fischer}{cs}
\end{icmlauthorlist}

\icmlaffiliation{cs}{Faculty of Computer Science, Ruhr University, Bochum, Germany}
\icmlaffiliation{math}{Faculty of Mathematics, Ruhr University, Bochum, Germany}

\icmlcorrespondingauthor{Mike Laszkiewicz}{mike.laszkiewicz@rub.de}

\icmlkeywords{Machine Learning, ICML}

\vskip 0.3in
]

\printAffiliationsAndNotice{\icmlEqualContribution} %

\begin{abstract}
In this work, we propose a set--membership inference attack for generative models using deep image watermarking techniques.
In particular, we demonstrate how conditional sampling from a generative model can reveal the watermark that was injected into parts of the training data. 
Our empirical results demonstrate that the proposed watermarking technique is a principled approach %
for detecting the 
non-consensual use of image data in training generative models.

\end{abstract}

\section{Introduction}
Recently developed generative models like StyleGAN \citep{karras2019style, karras2020analyzing, karras2021alias} and Stable Diffusion \citep{rombach2022high} enable users to generate new high-quality images in seconds.
In addition, the accessibility of code and computing resources has made it possible for users to easily fine-tune pre-trained generative models to adapt them to a certain style or subject \cite{hu2021lora, ruiz2022dreambooth} and publish them online.
However, this sparked some issues concerning data privacy and copyright \cite{forbes_getty, sd_litigation}.

Membership inference attacks (MIA) aim to determine whether certain data were used for training a machine learning model. 
In our case, a successful MIA can be useful to verify whether some copyright holder's images were used for training a generative model without consent.
This %
motivation indicates the importance of investigating (set) membership inference attacks on generative models in realistic deployment scenarios. 
While MIA has been tackled by prior works, they require either white-box access to the generative model or the use of shadow models~\cite{hayes2017logan, hilprecht_montecarlo, chen2020gan,  carlini2022membership, duan2023diffusion}. 
This is problematic since many generative models are hidden behind an UI or API that only returns the generated images (e.g. Midjourney, Dreamlike.art), making model parameters, internal activations, and other values that can be used in (semi-)white-box MIA techniques impossible to obtain. 
Moreover, shadow model training may be time- and cost-prohibitive.

The proposed set-MIA is based on contaminating the training data with watermarked samples. 
To confirm set--membership, we test whether the generative model reproduces the same watermark to a significant degree. This method is motivated by the observation that modern generative models are prone to replicate data~\cite{somepalli2022diffusion, carlini2023extracting} and recent advances in deep watermarking techniques~\cite{tancik2020stegastamp, yu2021artificial}. 
Our main findings based on an empirical evaluation on StyleGAN2 \cite{karras2020analyzing} are that (1) the generator does reproduce the watermark when trained on partially watermarked data but (2) reducing the proportion of watermarked training data strongly reduces the detectability of the watermark. However, (3) conditioning on synthetic samples similar to the watermarked images, substantially improves detectability.

\section{Attack scenario}
The adversary $\mathcal{C}$ in our setting is the model trainer who trains a generative model $G$, which maps $z \sim \mathcal{N}(0, I)$ to an image $G(z)=i\in \mathcal{I}$.
The adversary collects data $\mathcal{D}\subset \mathcal{I}$, which may include the data $\mathcal{P}\subset \mathcal{I}$.
The goal 
is to determine whether $G$ has been, in part or exclusively, trained on $\mathcal{P}$. 
For that, 
we have access to querying a set $\mathcal{T}$ of $n$ samples from $G$, without having any further knowledge about $G$. %
Querying from a generative model is typically done by accessing a UI or API provided by $\mathcal{C}$. Hence, to provide a set--membership inference tool with reasonable costs, we restrict the number of queries to $n=100$.

\section{Approach}
Our approach consists of injecting a detectable watermark on all $i \in \mathcal{P}$, such that, ideally, if $G$ is trained on $\mathcal{D} \supset \mathcal{P}$, it will generate samples with a similar detectable watermark. 
To accomplish this, we follow a deep watermarking approach~\cite{yu2021artificial}, where a watermark embedder $E :(\mathcal{I}, \{0, 1\}^{d}) \rightarrow \mathcal{I}$, and a watermark decoder $D: \mathcal{I} \rightarrow \{0, 1\}^{d}$ are trained together to embed a certain bit sequence $w \in \{0, 1\}^d$ into an image $i \in \mathcal{I}$ and decode it, respectively.
More formally, the optimization objective is given by 
\begin{equation*}\label{eq:obj}
    \arg \min_{E, D} \mathop{\mathbb{E}}_{\substack{i \sim \mathcal{I},\\ w\in \{0,1\}^d}} L_{\operatorname{BCE}}(w, D(E(i, w))) + \lambda L_{\operatorname{Rec}}(i, E(i, w)) , 
\end{equation*}
where $L_{\operatorname{BCE}}$ is a bitwise binary cross-entropy loss on the decoded watermark $D(E(i, w))$ 
and $L_{\operatorname{Rec}}$ is a reconstruction loss on the watermarked image $E(i, w)$.
After training $E$ and $D$, we can use $E$ to embed a watermark $w$ into every image in $\mathcal{P}$ to produce $\mathcal{P}':=\{ E(i, w): \; i \in \mathcal{P} \}$. 

MIA can then be performed by %
verifying whether $i \in \mathcal{T}$ possess the same watermark, for instance by measuring 
the number of correctly predicted bits $\operatorname{\#cor}(i):=\# \{j \in \{1, \dots, d\}: w_j = D(i)_j \}$, or equivalently, 
the bitwise accuracy $\operatorname{acc}(i):=\operatorname{\#cor}(i) / d$. 
Specifically, given $\mathcal{T}$, we can compute the one-sided $p$-value, which we denote as $p_{\operatorname{avg}}$, to test $H_0:  \operatorname{\#cor}(i) \vert w \sim Bin(d, p_w)$. 
Similarly, we can compute the $p$-value for the maximum bitwise accuracy over $\mathcal{T}$, $p_{\operatorname{max}}$, using the fact that under $H_0$ it is $\mathbb{P}(\max_{i \in \mathcal{T}} \operatorname{acc}(i) \geq \operatorname{acc}_{\operatorname{max}}) = 1- \prod_{i \in \mathcal{T}} \mathbb{P} (\operatorname{acc}(i) \leq \operatorname{acc}_{\operatorname{max}})$.
If the $p$-value is small, we have evidence that $G$ was most likely trained, partly, on $\mathcal{P}^\prime$. 
Note that in contrast to \citet{yu2021artificial}, we do not assume that $p_w=1/2$, because we have observed that $D$ might have a bias towards producing certain watermarks, see Table~\ref{table:uncond}. 
Instead, given a watermark $w$, we set $p_w$ as the average bitwise accuracy of $D(i)$ for non-watermarked real images $i$.

\section{Experiments}
In this section, we investigate how well $G$ reproduces the watermark when the watermarked data are diluted with non-watermarked data in the training set and how detection is affected by sampling images with the same observable attributes as $\mathcal{P}^\prime$.
We modify $E$ and $D$ slightly compared to \citet{yu2021artificial}: (1) we use the more modern ResNet blocks instead of stacking CNNs, (2) we learn to predict a small residual added to the original image and (3)  we use LPIPS~\cite{zhang2018unreasonable} as $L_{\operatorname{Rec}}$. 
We trained $(E, D)$ on CelebA, which comes with attribute labels for each sample. By embedding a watermark $w$ to each sample that has attribute $\frak{a}\in \{ \text{male}, \text{bushy eyebrows}, \text{eyeglasses} \}$, we end up with $3$ different $\mathcal{P}_{\frak{a}}^\prime$, which constitute $45.5\% $, $20.5\%$, $4.7\%$ of the training data $\mathcal{D}$, respectively. 
We trained a StyleGAN2 on $\mathcal{D} 
\supset \mathcal{P}_{\frak{a}}^\prime$ for each attribute $\frak{a}$.

\vspace{-0.15cm}
\paragraph{Detecting Unconditionally} 
Table~\ref{table:uncond} displays the bitwise watermark accuracies of $100$ unconditionally sampled images from the generative model. %
Unsurprisingly, the accuracy decreases as the size of $\mathcal{P}^\prime_{\frak{a}}$ is decreased. Nonetheless, given the null hypothesis that the watermarking accuracies are i.i.d. $Bin(100, p_w)$, we still end up with reasonably small $p$-values. Furthermore, based on the resulting $p$-values, we observe that the average accuracy is a more discriminative metric than the maximum accuracy. 

\begin{table}[t]
\caption{Bitwise watermark accuracies and corresponding $p$-values for varying  $\mathcal{P}_{\frak{a}}^\prime$. As a reference, we measure the bitwise accuracy obtained on the original CelebA dataset.
The extent to which we can recover $w$ scales negatively with the size of $\mathcal{P}^\prime_{\frak{a}}$. 
}
\label{table:uncond}
\vskip 0.1in
\begin{center}
\begin{small}
\begin{sc}
\begin{tabular}{lcccc}
\toprule
$\frak{a}$ & $\operatorname{acc}_{\operatorname{avg}}$ & $p_{\operatorname{avg}}$ & $\operatorname{acc}_{\operatorname{max}}$ & $p_{\operatorname{max}}$\\
\midrule
Male   & $69.04\%$  & $7.5e^{-271}$  & $100.0\%$ & $0.00$  \\
Eyebrows &  $54.91\%$  & $1.0e^{-10}$  & $68.00 \%$ & $0.03$ \\
Eyeglasses    & $52.65\%$  & $0.03$  & $67.00 \%$ & $0.07$  \\
\midrule 
CelebA   & $51.74 \%$  & $0.50$  & $60.00 \%$ & $0.98$  \\
\bottomrule
\end{tabular}
\end{sc}
\end{small}
\end{center}
\vskip -0.1in
\end{table}

\vspace{-0.15cm}
\paragraph{Detecting Conditionally} %
Instead of sampling unconditionally, we propose to choose samples from $G$ conditioned on $\frak{a}$ to infer the membership of $\mathcal{P}_{\frak{a}}'$. 
Since we generally cannot generate conditionally, we instead use an attribution predictor\footnote{\url{https://github.com/d-li14/face-attribute-prediction}} to select a subset $\mathcal{T}_{\frak{a}}$ of $\mathcal{T}$ according to $\frak{a}$. 
Again, we measure the bitwise accuracies of the inferred watermarks, shown in Table~\ref{table:cond}.
Note that for each $\frak{a}$, we set $p_w$ as the average bitwise accuracy achieved for CelebA samples possessing $\frak{a}$.
Even though the effective sample size of $\mathcal{T}_{\frak{a}}$ is reduced, we observe a clear increase in the statistical significance as illustrated by the $p$-values, compared to the unconditional case.  
\begin{table}[t]
\caption{Bitwise watermark accuracies and corresponding $p$-values when conditioning on $\frak{a}$. We can recover $w$ up to a significant amount, which gives clear evidence for the membership of $\mathcal{P}_{\frak{a}}^\prime$.  
}
\label{table:cond}
\vskip 0.1in
\begin{center}
\begin{small}
\begin{sc}
\begin{tabular}{lcccc}
\toprule
$\frak{a}$ & $\operatorname{acc}_{\operatorname{avg}}$ & $p_{\operatorname{avg}}$ & $\operatorname{acc}_{\operatorname{max}}$ & $p_{\operatorname{max}}$\\
\midrule
Male   & $93.66\%$  & $0.00$  & $100.0\%$ & $0.00$  \\
Eyebrows &  $64.33\%$  & $9.8e^{-19}$  & $68.00 \%$ & $4.1e^{-3}$ \\
Eyeglasses    & $60.83\%$  & $5.1e^{-6}$ & $67.00 \%$ & $4.4e^{-3}$  \\
\bottomrule
\end{tabular}
\end{sc}
\end{small}
\end{center}
\vskip -0.1in
\end{table}

\section{Conclusion and Future Work}
In this paper, we propose a set--membership inference attack for generative models based on embedding an invisible watermark in parts of the training data. 
The provided experiments demonstrate that generative models create samples that possess this injected watermark, which can be used to prove set--membership. 
Moreover, we propose to use generated samples conditioned to be similar to the watermarked training data 
because we observed that they reproduce the watermark significantly better.
A promising direction for future work is extending this analysis to modern text-to-image models, which are conditional models by \mbox{construction}.
\bibliography{Literatur}
\bibliographystyle{icml2023}

\end{document}